\documentclass[11pt]{article}

\usepackage[utf8]{inputenc} % allow utf-8 input
\usepackage[T1]{fontenc}    % use 8-bit T1 fonts
\usepackage{hyperref}       % hyperlinks
\usepackage{url}            % simple URL typesetting
\usepackage{booktabs}       % professional-quality tables
\usepackage{amsfonts}       % blackboard math symbols
\usepackage{nicefrac}       % compact symbols for 1/2, etc.
\usepackage{microtype}      % microtypography
\usepackage{lipsum}
\usepackage{array,multirow}
\usepackage{blindtext}
\usepackage{graphicx}
\usepackage{xcolor}
\usepackage[cmex10]{amsmath}
\usepackage{amssymb}
\usepackage{amsthm}
\usepackage[tight,footnotesize]{subfigure}
\usepackage{caption}
\usepackage{booktabs}
\usepackage{comment}
\usepackage{algorithm}
\usepackage{algorithmic}

\newcommand{\bR}{\mathbb{R}}
\newcommand{\bC}{\mathbb{C}}
\newcommand{\bS}{\mathbb{S}}
\newcommand{\bP}{\mathbb{P}}

\newcommand{\sL}{\mathcal{L}}

\newtheorem{theorem}{Theorem}

\newtheorem{example}{Example}
\newtheorem{definition}{Definition}

\begin{document}

\title{Machine learning the real discriminant locus}

\author{
Edgar A. Bernal\thanks{FLX AI, Rochester, NY 14607 (edgar.bernal@flxai.com).}
\and
Jonathan D.~Hauenstein\thanks{Department of Applied and Computational Mathematics and Statistics,
University of Notre Dame, Notre Dame, IN 46556 (hauenstein@nd.edu, \url{www.nd.edu/\~jhauenst}).
This author was supported in part by
NSF grant CCF-1812746 and
ONR N00014-16-1-2722.}
\and
Dhagash Mehta\thanks{The Vanguard Group, Malvern, PA 19355 (dhagashbmehta@gmail.com).}
\and
 Margaret H.~Regan\thanks{Department of Mathematics,
Duke University, Durham, NC 27708 (mregan9@math.duke.edu,
\url{www.margaretregan.com}).
This author was supported in part by
Schmitt Leadership Fellowship in Science and Engineering and NSF grant CCF-1812746.}
\and Tingting Tang\thanks{Department of Mathematics and Statistics, San Diego State University, Imperial Valley, CA, 92231.}
}

\maketitle

\begin{abstract}
\noindent
Parameterized systems of polynomial equations arise in many
applications in science and engineering with the real solutions describing, for example, 
equilibria of a dynamical system, linkages satisfying design constraints, 
and scene reconstruction in computer vision.  
Since different parameter values can have a different number of real solutions,
the parameter space is decomposed into regions whose boundary forms the 
real discriminant locus.
This article views locating the real discriminant locus as a 
supervised classification problem in machine learning
where the goal is to determine classification boundaries over the parameter space, with the classes being the number of real solutions.
This article presents a novel sampling method which carefully samples a
multidimensional parameter space.
At each sample point, homotopy continuation is used to  obtain the number of real solutions to the corresponding polynomial system.
Machine learning techniques including nearest neighbors, support vector classifiers, and neural networks are used to efficiently approximate the real discriminant locus.
One application of having learned the real discriminant locus
is to develop a real homotopy method that only tracks real solution paths
unlike traditional methods which track all complex solution paths. Examples show 
that the proposed approach 
can efficiently approximate complicated solution boundaries such as those arising from 
the equilibria of the $N=4$ Kuramoto model which was previously intractable using traditional methods.
\end{abstract}


\section{Introduction}

Systems of polynomial equations are a collection of multivariate nonlinear equations in which each equation is a multivariate polynomial. Such systems arise naturally in many areas of science and engineering ranging from 
chemistry, particle physics, string theory, mathematical biology, phylogenetics, control theory, robotics, power systems, and computer vision \cite{BHSW13, CLO:98, CLO:07, Cox20, SW05}. 
In many of these applications, the coefficients of the equations depend upon one or more parameters
yielding parameterized systems of polynomial equations.
Both the solutions and the number of real solutions are
functions of the parameters.  Investigating the solution structure 
as a function of the parameters is typically more difficult than solving the system for given values of~the~parameters.

Due to its ubiquity, many methods
have been proposed to characterize the solution structure
over the parameter space.  Classically, the discriminant
describes the boundary between regions where the solution structure
changes \cite{GKZ}.  For example, the discriminant of
\begin{equation}\label{eq:Quadratic}
f(x;b,c) = x^2 + bx + c
\end{equation}
is $D = b^2-4c$.
Since real parameters and the number of real solutions are of most interest 
in applications
describing, for example, 
equilibria of dynamical systems
and 
scene reconstruction in computer vision,
this paper focuses 
on the boundary between regions where the number of real solutions change.  
Thus, from this perspective, we 
concentrate on the real discriminant locus
associated with counting the number of real solutions.

\begin{definition}\label{def:RDL}
Given a parameteric polynomial system $f(x;p): \mathbb{R}^n\times\mathbb{R}^k\rightarrow\mathbb{R}^n$, its real discriminant locus is the boundary in the parameter space where the number of real solutions changes.  
\end{definition}

For~Eq.~\eqref{eq:Quadratic}, the solution set in~$\bR^2$
of $D=0$ is the real discriminant locus, which forms the boundary
between the region in $\bR^2$ with $D>0$ where $f = 0$ has two real solutions
and the region in $\bR^2$ with $D<0$ where $f=0$ has no real~solutions.
In addition to just the number of real solutions, one could also be interested
in additional structure related to the real solutions, e.g., geometric properties, which would then define a corresponding real discriminant locus.  See \cite{DBLP:LazardR07} for more details. 

%of the real solutions.  This additional structure would then define a corresponding real discriminant locus.  See \cite{DBLP:LazardR07} for more details. }

Comprehensive Gr\"obner basis computations \cite{Weis:92}
can be used to symbolically compute the discriminant polynomial over the complex numbers
whose solution set, i.e., algebraic variety, is called the
complex discriminant locus.
Since the discriminant actually defines the boundaries over the complex
parameter space, one can develop specialized methods over the real numbers.
Some examples include
cylindrical algebraic decomposition~ \cite{bpr:03,hanan2010stability,hernandez2011towards,DBLP:LazardR07,xia2007discoverer},
Brouwer degree \cite{conradi2017identifying}, and polyhedral methods \cite{bihan2018lower,giaroli2019regions} 
which have been utilized for modest size systems.  However, computational methods 
which depend upon Gr\"obner basis or other symbolic computations severely
suffer from exponential complexity.

To mitigate these issues, global symbolic methods can be replaced by local, numerical approximations to determine the discriminant locus.  For larger systems, numerical methods based on a form
of homotopy continuation \cite{allgower2012numerical,BHSW13,SW05} have been employed
in which one tracks the solution structure as the parameters
are varied continuously.  
Several computational packages such as AUTO~\cite{doedel1981auto} and MATCONT \cite{dhooge2003matcont} employ such techniques
for parameterized differential equations.  
Rather than directly run into the real discriminant locus, a perturbed
sweeping approach was presented in \cite{harrington2016decomposing}.
In particular, when all complex solutions over a general parameter point can be computed,
homotopy continuation provides an approach to obtain global information
about the solutions which, for example, can be used to obtain
the number of real solutions at selected sample points
in the parameter space
\cite{Paramotopy,chandra2017locating,greene2013tumbling,he2013exploring,MartinezPedrera:2012rs}.  

Our method aims to numerically approximate the real discriminant locus by viewing this as a classification problem in machine learning. The problem of approximating the
real discriminant locus is posed 
as a problem of approximating the decision boundaries that separate data points 
according to their labels, where the input features 
are the parameters of the given parametric system and the target labels are the number of real solutions at the corresponding parameter values.  Given a parameter point, homotopy continuation can be used to generate
the labels, i.e., compute the number of real solutions.  A novel approach
for selecting sample points is developed 
by leveraging domain knowledge obtained 
from numerical algebraic geometry \cite{BHSW13,SW05}
to help guide the approximation of the decision boundaries.

Although there has been a collection of papers,
such as \cite{Das2012,huang2002constrained,huang2004constructive,huang2001neural,huang2004neural,huang2018using,perantonis1998constrained},
attempting to solve polynomial equations and improving algorithms using
neural networks, the approach closest to the present work is~\cite{mourrain2006determining}.
That work uses a feed-forward neural network with one hidden layer to predict the 
number of real solutions for univariate polynomials. 
All the sample points, including training data and testing data are combinations of 
integer coefficients in the parameter space. 
The results indicate that the ability of an artificial neural network to generalize 
on the test sets is comparable to its performance on the training sets. 
Employing several algorithms to train the network for high degree polynomials showed that the choice of training program impacts the performance of the network.

Other applications of deep networks for analyzing polynomial equations
include \cite{andoni2014learningNeuralNet,andoni2014learning}
which investigated the effectiveness of deep networks to learn
a target function that is a low degree polynomial.
A neural network which is used as a pre-training method for finding the 
number of real solutions and then used to design a neural network-like model to 
compute real solutions to univariate polynomial in parallel is provided in \cite{Das2012}.
Finally, \cite{breiding2018learning} employed machine learning algorithms to 
learn the geometry and topology of the complex solution set of systems of 
polynomial equations.

This manuscript provides a novel approach to analyzing the 
real solution structure of parameterized polynomial equations
which are multivariate and depend upon many parameters.
The specific contributions are as follows:
(1) Transform the problem of computing the real discriminant locus of 
    parameterized polynomial equations into a supervised classification problem in order to use machine learning constructs such as nearest neighbor, support vector classifiers, and deep learning techniques; (2) Devise a novel sampling technique that leverages domain knowledge from
    numerical algebraic geometry which can be thought of as a static active learning implementation where the desired training set is determined in advance;
(3) Show that machine learning techniques can quickly approximate the real discriminant locus even when they contain cusps and other singular regions which, in turn, provides a decomposition of the parameter space into regions where the number of real solutions remains constant; (4) Demonstrate that the proposed method was able to break a ceiling in computational algebraic geometry
as the new approach is able to analyze the $N=4$ Kuramoto model which was previously intractable 
using traditional methods; (5) Design a real homotopy method that utilizes an approximation of the real discriminant locus to track only real solution paths and computes only real solutions, thus improving efficiency.

The rest of the paper is organized as follows.
Section~\ref{sec:computational_methods}
provides background information on numerical algebraic geometry, 
homotopy continuation, and parameterized systems.
Similarly, Section~\ref{sec:machine_learning}
provides an overview of the machine learning techniques 
utilized: nearest neighbor, 
support vector classifiers,
and deep learning.
The novel sampling scheme 
that leverages domain knowledge from numerical algebraic geometry is presented in Section~\ref{sec:sampling}.
Section~\ref{sec:results} applies the proposed approach to several examples. A real homotopy method is outlined in 
Section~\ref{sec:realHomotopy}, which utilizes the learned real discriminant locus
to track only real solution paths.
The paper concludes in~Section~\ref{sec:conclusion}.

\section{Numerical algebraic geometry}\label{sec:computational_methods}

The following describes 
parameter homotopies and pseudowitness point sets,
which are two topics in numerical algebraic geometry
that will be used to learn the real discriminant locus.
See \cite{BHSW13,SW05}
for more details regarding numerical algebraic geometry.

\subsection{Parameter homotopies}\label{sec:parameterHomotopies}

For simplicity, we consider parameterized polynomial systems of the form
\begin{equation}\label{eq:parameterSystem}
f(x;p) = f(x_1,\dots,x_n;p_1,\dots,p_k) = 
\hbox{\scriptsize $
\left[\begin{array}{c} f_1(x_1,\dots,x_n;p_1,\dots,p_k) \\
\vdots \\ f_n(x_1,\dots,x_n;p_1,\dots,p_k) 
\end{array}\right]$}
\end{equation}
such that, for a generic $p^*\in\bC^k$, 
the algebraic variety defined by
$f(x;p^*) = 0$ 
consists of finitely many points
in $\bC^n$, say $d$,
all of which are nonsingular.  A solution $x^*$ of $f(x;p^*)=0$
is nonsingular if $J_x f(x^*;p^*)$ is invertible where
$J_x f(x;p)$ is the Jacobian matrix of $f$ with respect to $x$.
This is the typical situation for  well-constrained parameterized polynomial systems arising in science and engineering
applications.
For reducing overdetermined parameterized systems to well-constrained parameterized systems adhering to Eq.~\eqref{eq:parameterSystem}, 
see \cite{HauensteinRegan} and the references therein.  
One can also reduce 
to the nonsingular isolated
case for parameterized systems which generically
define a positive-dimensional algebraic
variety using linear slicing
and singular isolated solutions using deflation, e.g., see \cite{Isosingular,Deflation}.

A key consequence of this setup is that
the real parameter space $\bR^k$ contains open subsets where the number
of real solutions to $f(x;p) = 0$ is constant.  The boundaries 
of these open subsets form the {\em real discriminant locus}.  

\begin{example}\label{ex:Quadratic}
The parameterized quadratic described by Eq.~\eqref{eq:Quadratic} 
generically has $d=2$ solutions in~$\bC$.  
For $D = b^2-4c$, the two open subsets defined by
$D>0$ and $D<0$ in $\bR^2$ have a constant 
number of real solutions, namely~$2$ and $0$.  The real discriminant 
locus is the algebraic variety defined by $D=0$ in $\bR^2$.  
\end{example}

The complex discriminant locus consists of the parameter points in $\bC^p$
where $f(x;p)=0$ does not have $d$ nonsingular solutions.
Since $f(x;p)$ is well-constrained, the complex discriminant locus
is either empty or is a hypersurface in $\bC^p$.  
Section~\ref{sec:sampling} 
exploits this fact to generate sample
points on the real discriminant locus, which is contained in the complex discriminant locus.
The following illustrates this
while showing that the real discriminant locus
could have~smaller~dimension.  

\begin{example}\label{ex:ParameterDisc}
Consider the following from \cite[Ex~2.1]{RealMonodromy}:
$$
\hbox{\small $
f(x;p) = \left[\begin{array}{c}
x_1^2-x_2^2-p_1 \\ 2x_1x_2-p_2
\end{array}\right]$.}$$
The system of equations $f(x;p) = 0$ 
generically has $d=4$ solutions
with $p_1^2+p_2^2=0$ in $\bC^2$ defining the 
complex
discriminant locus.  Thus, the complex discriminant locus is a curve in $\bC^2$,
while the only point in~$\bR^2$ on this complex hypersurface is the origin.  
Moreover, for all $p\in\bR^2\setminus\{(0,0)\}$,
$f(x;p) = 0$ has~$2$ real solutions showing that the real discriminant locus in $\bR^2$ 
is $\{(0,0)\}$.
\end{example}

Given $p\in\bC^k$ outside of the complex discriminant locus, 
the solutions to $f(x;p) = 0$
can be computed using a parameter homotopy \cite{morgan1989coefficient}.
In order to utilize a parameter homotopy,
one first needs to know 
a parameter value $p^*\in\bC^k$ and a set
$S\subset\bC^n$ consisting of the $d$
solutions to $f(x;p^*) = 0$. 
Obtaining this starting information is
called the {\em ab initio}
phase.
See \cite[Chap.~6]{BHSW13} for more details.

\paragraph{Ab initio phase}
The input for a parameter homotopy
is a parameter value $p^*\in\bC^k$ and 
the algebraic variety $S\subset\bC^n$ 
defined by $f(x;p^*) = 0$ consisting of $d$ points.
One can compute $S$ by using a 
classical linear homotopy.
For example, if $e_i = \deg f_i$
and $g_i(x) = x_i^{e_i}-1$, then 
$$\hbox{\small $
H(x,t) = (1-t)f(x;p^*) + \gamma t g(x) = 0$}$$
where $\gamma\in\bC$ is generic
is called a total degree homotopy consisting of $\prod_{i=1}^n e_i$ paths.
Clearly, the solutions of $H(x,1) = g(x)=0$ are trivial to compute providing the
$\prod_{i=1}^n e_i$ start points.
For $t\in(0,1]$, the solution paths defined
by $H(x,t) = 0$ are smooth and can be 
traversed using a variety of numerical methods \cite{PathTrackMethods,BHSW13}.  
The $d$
solution paths which have a finite limit as~$t$ approaches $0$ converge to the $d$
solutions of $f(x;p^*)=0$.  
By assumption on the structure of $f$,
the other solution paths will diverge
to infinity.  To possibly reduce the number
of paths that diverge, one can select
a different structure for $g(x)$
such as based on the multihomogeneous structure 
of $f$ or the monomial structure of $f$, e.g., see 
\cite[Chap.~8]{SW05}.

\paragraph{Parameter homotopy phase}
With the {\em ab initio}
phase complete, the ``online''
parameter homotopy phase can commence
to solve $f(x;p) = 0$ for various $p\in\bC^k$.
One utilizes the parameter homotopy
\begin{equation}\label{eq:ParameterHomotopy}
\hbox{\small $
H(x,t) = f(x;\tau(t)\cdot p^* + (1-\tau(t))\cdot p) = 0$}
\hbox{~~~~where~~~~}
\hbox{\small $\tau(t) = \dfrac{\gamma t}{1+(\gamma-1)t}$}
\end{equation}
such that $t\in[0,1]$ and $\gamma\in\bC$.
In particular, $H(x,1) = f(x;p^*)=0$ has known solutions $S$,
computed in the {\em ab initio}
phase,
and one aims to compute the solutions to $H(x,0) = f(x;p)=0$.  
For generic values of the constant $\gamma\in\bC$, the 
arc $\tau(t)\cdot p^* + (1-\tau(t))\cdot p$ 
for $t\in[0,1]$ connects $p^*$ to $p$ 
and avoids the complex discriminant locus.
Thus, for $t\in[0,1]$,
$H(x,t) =0$ defines 
precisely $d$ solution paths 
connecting the~$d$~points in $S$
with the $d$ solutions to $f(x;p)=0$.
As above, numerical methods
can be used to track the paths
and a certified count on the number of real 
and nonreal solutions can be obtained,
e.g., see \cite{alphaCertified}.

When $p\in\bR^k$, the number of complex solutions $d$ can be significantly larger than the 
number of real solutions to $f(x;p)=0$.
Thus, Section~\ref{sec:realHomotopy} considers a real parameter homotopy 
aiming to only track real solution paths by trying to stay within
each open subset of the parameter space where the number of real solutions~is~constant.
In particular, if the real discriminant locus has smaller dimension,
such as in Ex.~\ref{ex:ParameterDisc}, this is beneficial since it becomes
easier to avoid intersecting the real discriminant locus.  
Therefore, our learning of the 
real discriminant locus in Section~\ref{sec:machine_learning}
and sampling scheme in Section~\ref{sec:sampling} 
is only concerned with the codimension $1$ boundaries in~$\bR^k$.

\subsection{Pseudowitness point sets}\label{sec:WitnessSets}

The key to the sampling method in Section~\ref{sec:sampling} 
is to utilize domain knowledge from the complex discriminant locus
to select sample points to guide the learning of the real discriminant locus.
Rather than computing a polynomial defining the complex discriminant locus,
which can often be a computationally challenging problem, 
the method in Section~\ref{sec:sampling} computes a
pseudowitness point set~\cite{HSprojection,HSmembership}
by intersecting 
the complex discriminant locus
with a real line.
This permits one to perform geometric
computations on the complex discriminant
locus without explicitly needing
to compute its defining equation.

When all $d$ solution paths remain finite
when performing a
parameter homotopy using Eq.~\eqref{eq:ParameterHomotopy} for 
every $p\in\bC^k$,
then one can compute a pseudowitness point set for
the complex discriminant locus as follows.
For $f(x;p)$ as in Eq.~\eqref{eq:parameterSystem}, consider
the system 
$$
\hbox{\small $
F(x,p) = \left[\begin{array}{c} f(x;p) \\ \det J_x f(x;p) \end{array}\right]$.}$$
Let $V\subset\bC^{n+k}$ be the 
algebraic variety defined by
$F(x,p) = 0$,
$\pi(x,p) = p$
be the projection map onto the parameters, 
and $\sL\subset\bC^k$ be a general line.
Then, the pseudowitness point set for $V$ with respect to 
the projection map $\pi$ and line $\sL$
is~$\pi(V)\cap\sL$.  
The number of points in $\pi(V)\cap\sL$
is the degree of the complex discriminant locus.
One can treat the coefficients
of the line~$\sL$ as parameters
and utilize a parameter homotopy to 
deform the line $\sL$
to compute a pseudowitness point set 
corresponding to
other lines in $\bC^k$.
Hence, this provides a method for sampling points on the complex discriminant locus. 
Note that a null space
approach $J_x f(x;p)\cdot w$ for $w\in\bP^{n-1}$
may be used 
instead of $\det J_x f(x;p)$ \cite{BHPS10}.

If some solution paths of a parameter homotopy diverge to infinity, then one can projectivize the variables $x$
to compactify the fiber over each param\-eter point $p$, e.g.,
see~\cite[\S~3]{HSmembership},
and then~proceed~as~above.

\section{Machine learning}\label{sec:machine_learning}

Parameter homotopies discussed in Section~\ref{sec:parameterHomotopies}
provide a means for counting the number of real solutions
corresponding to a given parameter value.  
Indeed, there are other options 
such as 
using Hermite matrices~\cite{SzantoHermite,SafeyElDinHermite}.  Machine learning techniques
can use the number of real solutions as
labels and make predictions
about previously unseen parameter points.  
This setup follows a supervised learning paradigm in machine learning 
since the labels are known for training data.
Moreover, approximating the real discriminant locus is equivalent to approximating the decision boundaries between different classes. 
Since Section~\ref{sec:realHomotopy}
applies the learned boundaries
to construct a real parameter homotopy
which requires knowing the real solutions
rather than just the number of them, 
we utilize parameter homotopies in our computations.

The following describes 
the leveraged machine learning techniques, namely $k$-nearest neighbors ($k$-NN), support vector classifiers (SVC) and feedforward neural networks.

\subsection{\texorpdfstring{$k$}{k}-nearest neighbors}\label{subsec:knn}

The underlying premise of a nearest neighbor classification algorithm
is that the class to which a previously unseen data sample belongs
can be inferred from the class to which the most similar samples in the 
training set belong.  
In our context, similarity will be measured in the form of the Euclidean distance
using $k$ samples in the training set nearest to the test sample
thereby yielding the $k$-nearest neighbors.  
The label assigned to the previously unseen data sample
is simply the class to which to the majority of the $k$-nearest neighbors belong.

In addition to being easy to implement, a $1$-nearest neighbor 
classification algorithm has desirable properties to our problem.  
In particular, the Bayes error rate is the lowest misclassification rate achievable 
by any classifier on the associated data~\cite{Fukunaga:1990:ISP:92131,546912}.  
Since the labels are deterministic and the classes do not overlap for our problem, the Bayes error rate is equal to 0. 
This is summarized in the following.

\begin{theorem}\label{thm:1NN}
Provided the parameter space is sampled densely enough, 
no other classifier will outperform a $1$-nearest neighbor 
classification algorithm for determining
the number of real solutions associated with a given parameter~point. 
\end{theorem}
\begin{proof}
The result follows from the fact that, as the number of training samples tends to infinity, 
the error rate of any given classifier is at worst its 
Bayes error rate~\cite{1053964,Ripley:1995:PRN:546466}
with the best possible error rate attainable by any classifier being 0.
Since, in this case, the Bayes error rate is indeed 0 due to the non-overlapping nature of the classes, 
no other classifier can possibly improve upon the asymptotic behavior of the 1-nearest neighbor classifier.
\end{proof}

Clearly, Theorem~\ref{thm:1NN} has significant practical limitations since both the 
complexity and the storage requirements of naive implementations, i.e., non-tree-based methods, 
for a $1$-nearest neighbor classification algorithm
are $\mathcal{O}(k\ell)$ when the parameter space is $\bR^k$
and $\ell$ is the cardinality of the training set~\cite{Weber:1998:QAP:645924.671192}.
Therefore, implementing a truly optimal version would be unfeasible. 
One approach to partially overcome these strict computational requirements 
is by implementing a sampling technique that utilizes
domain knowledge as described in Section~\ref{sec:sampling} 
which can be viewed as a form of selective sampling~\cite{pmlr-v23-dasgupta12,Lindenbaum:1999:SSN:315149.315323}, a type of active learning~\cite{Aggarwal_chapter22,settles2009active}.
This enables us to ameliorate the impact of the trade-off between the number of samples 
stored and algorithmic~performance.

The techniques used in $k$-nearest neighbor algorithms belong to memory-based classification methods as they require the entire training set to be stored. 
The next two subsections discuss sparse kernel methods, which classify new inputs based on computations performed on a subset of the training set, followed by parametric methods, which learn a set of parametric classification rules based on the training data and perform decision-making based on the learned rules only without referring back to training samples.

\subsection{Support vector classifiers}
Perhaps the most popular instance of so-called kernel methods are support vector classifiers \cite{christopher2006pattern}. The term kernel refers to a (typically) nonlinear mapping that is effected on the training data points, and classification is performed in the resulting nonlinear space. Kernel mapping has both computational and capacity-related advantages since it enables reasoning in a high-dimensional space (usually higher-dimensional than the original feature space) without explicitly computing the high-dimensional representation of data points. Rather, only inner products between kernel representations of the samples are involved \cite{Theodoridis:1314938}.  
Inter-class boundaries for SVCs are computed by maximizing the gap between the samples in the different classes.

In real-life scenarios, where it may be difficult to find a representation space in which classes are separable, overfitting less representative samples in the training data, in particular those that cross inter-class boundaries, typically results in poor generalization abilities of the network to classify unseen data.  Since this is usually associated with excessive network capacity, regularization techniques are often implemented~\cite{Goodfellow-et-al-2016}.  Commonly used regularization techniques include $L_1$ (Lasso) and $L_2$ (Ridge) regularization, dropout, and early stopping \cite{bengio2015deep,bishop2006pattern}.  We adopt a strategy that goes against this widely accepted principle.  
The reason for this is that we know {\em a priori} that the 
training data originated from counting the number of real solutions to a 
parameterized system of polynomial equations, which can be certifiably computed
as discussed in Section~\ref{sec:parameterHomotopies}.  
The benefit of knowing the provenance of the data is the awareness that the data in question is separable. Therefore, in order to closely approximate the underlying structure of the training data, which closely follows the decision boundaries without inter-class overlap, we deliberately minimize the degree of regularization in our models. In the context of SVCs, this is achieved in practice by choosing a large inverse regularization parameter $c$.

\subsection{Neural networks}

Backed by the universal approximation theorem~\cite{cybenko1989approximation,hornik1989multilayer},
deep learning techniques \cite{bengio2015deep,lecun2015deep} have garnered significant 
popularity in recent times based on success in a wide array of applications.
In particular, the feedforward neural network, i.e., a multi-layer structure of 
compositions of activation functions, has been shown to be a universal approximator for any 
mildly-constrained target function provided that the network parameters (or weights) and the multilayer structure are chosen appropriately~\cite{cybenko1989approximation,hornik1989multilayer}. 
The layers of compositions of functions manifest the multilayer structure in a network where the depth refers to the number of composition levels.

A practical way to obtain a sensible model and its corresponding weights is to start with a large architecture (as a rule of thumb, as many weights as the number of training data points)
and apply an optimization routine, e.g., stochastic gradient descent method, 
to achieve numerical values of the weights which best approximate the underlying function. As motivated above, we completely forego the use of regularization techniques in the training process of our neural networks.

\section{Sampling method}\label{sec:sampling}

Given a compact subset of the parameter space $\bR^k$, one approach to generate sample points is to 
randomly, e.g., uniformly, select a parameter value and use a parameter homotopy
to count the number of real solutions.  
Such an approach has been applied 
to a variety of problems, e.g.,~\cite{Paramotopy,alphaCertified,ParameterGeography}.
Aiming to approximate the real discriminant
locus (classification boundaries),
the following uses domain knowledge 
via the complex discriminant locus to find
sample points near the boundaries to guide
the learning~of~the~boundaries.

For a parameterized polynomial system $f(x;p)$
as in Eq.~\eqref{eq:parameterSystem}, we start
with parameter homotopies for solving $f=0$ (see Section~\ref{sec:parameterHomotopies})
and computing a pseudowitness point set
for the complex discriminant locus (see Section~\ref{sec:WitnessSets}).
The sampling method
starts with a randomly
selected parameter value $p^*\in\bR^k$,
e.g., uniformly sampled in a compact subset $\Omega$ 
of the parameter space $\bR^k$.
For simplicity, we assume that $\Omega$ is a rectangular
box.  The parameter homotopy for $f=0$
is used to count the number of real solutions
to $f(x;p^*)=0$ thereby obtaining the label for $p^*$.

The key addition in the sampling scheme is to then
select a random direction $v^*$ uniformly in~$\bS^{k-1}$,
the unit sphere in $\bR^k$.  
Let $\sL^*\subset\bC^k$ be the line parameterized
by $p^*+\lambda\cdot v^*$ for $\lambda\in\bC$.  Then,  the parameter homotopy for computing a pseudowitness 
point set for the complex discriminant locus
is used to compute the real points 
in the corresponding pseudowitness point set along $\sL^*$
inside of~$\Omega$,
say $p_1=p^*+\lambda_1\cdot v^*,\dots,p_\ell=p^*+\lambda_\ell\cdot v^*$.
Without loss of generality, we can assume $\lambda_1<\lambda_2<\cdots<\lambda_\ell$.
Compute $\lambda_0$ and $\lambda_{\ell+1}$ 
such that $\lambda_0 < \lambda_1 < \lambda_\ell < \lambda_{\ell+1}$ where
$p_0=p^*+\lambda_0\cdot v$ and $p_{\ell+1}=p^*+\lambda_{\ell+1}\cdot v$
are the intersection points of $\sL^*$ with 
the boundary of $\Omega$.  
We note that if one has access to the 
complex discriminant polynomial~$D$,
e.g., using \cite{Harris2020,SafeyElDinHermite,Weis:92},
then an alternative to compute $\lambda_1,\dots,\lambda_\ell$ would be 
via computing 
real roots of the univariate polynomial 
$D(p^*+\lambda\cdot v^*)$.  

Along $\sL^*$, the complex discriminant locus
yields that the number of real solutions
is constant on the intervals $(p_i,p_{i+1})$
contained in $\sL^*$ for $i=0,\dots,\ell$.
Hence, the next step
is to determine the number of real solutions 
associated with each interval $(p_i,p_{i+1})$.
This is accomplished by 
selecting the midpoint of each interval,
namely $m_i = p^*+(\lambda_i+\delta_i/2)\cdot v$ 
for $i=0,\dots,\ell$ 
and $\delta_{i}=\lambda_{i+1}-\lambda_i$.
The parameter homotopy for $f=0$ is used
to count the number of real solutions of $f(x;m_i)=0$.

Our sampling scheme takes the midpoints $m_i$
of each interval, which we call ``near center'' points
in the corresponding cell.  We add
``near boundary'' points as follows.
Given $\alpha > 0$, the near boundary points
are $b_{i,f} = p^*+(\lambda_i+\Delta_i^f)\cdot v$
and $b_{i,b} = p^*+(\lambda_i-\Delta_i^b)\cdot v$
for $i=1,\dots,\ell$
where $\Delta_i^f = \min\{\alpha,\delta_i/20\}$
and $\Delta_i^b = \min\{\alpha,\delta_{i-1}/20\}$.
Since $b_{i,f}\in(p_i,p_{i+1})$
and $b_{i,b}\in(p_{i-1},p_i)$, the number of real solutions
of $f(x;b_{i,f})=0$ 
and $f(x;b_{i,b})=0$ are known from the 
computation~above.

The aim of the near center points is
to provide a parameter point sufficiently 
in the interior of the region in $\bR^k$
with the same number of real solutions.
The aim of the near boundary points is
to help learn the boundary by providing
points on either side of the boundary.
Of course, one could also explicitly 
force the learned boundary to pass
through the sampled boundary points.  
However, they are not utilized in
Section~\ref{sec:results} since the
near boundary points provide both interior
points of the corresponding regions 
as well as guide the learning of the boundary.

\begin{figure}[!t]
    \centering
    \includegraphics[scale = 0.35]{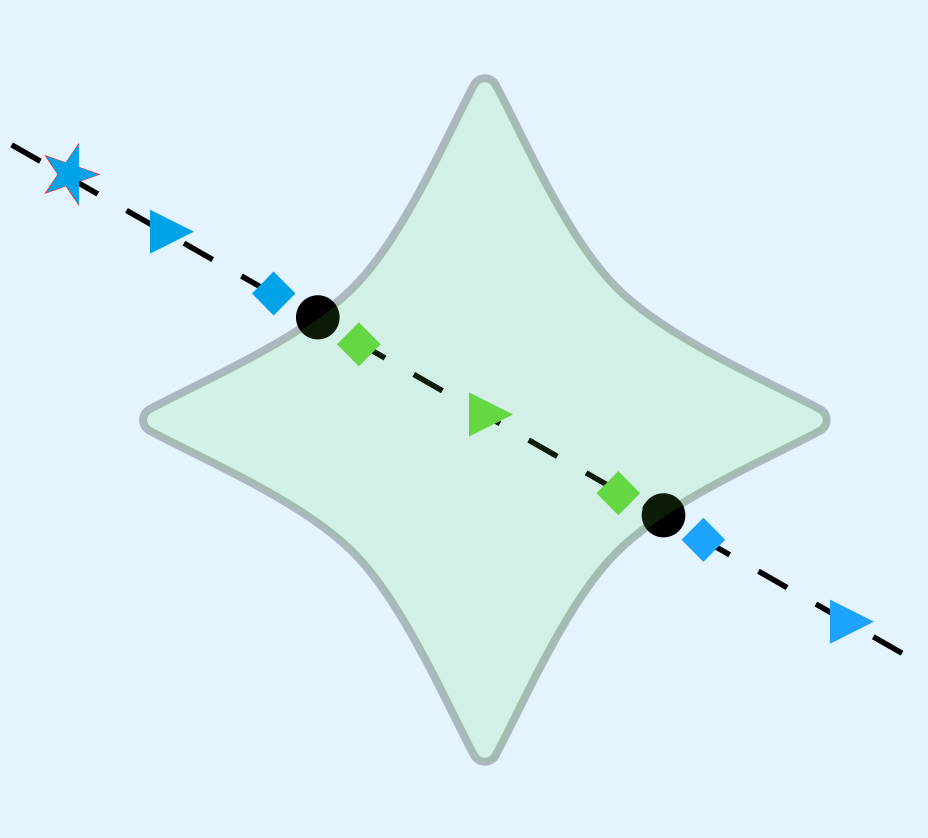} 
    \caption{Visual representation of the sampling scheme where the star is a uniform random sample point, circles are points on the boundary, triangles are midpoints, and 
diamonds are near boundary points.  
The points are color coded based on the number of real solutions.}
    \label{fig:sampling}
\end{figure}

In total, our sampling scheme utilized
in Section~\ref{sec:results} 
provides three different
types of data points: uniform points, near center points, and near boundary points. 
Figure~\ref{fig:sampling} provides an illustration
of these point categories based on a selected uniformly 
selected sample point (star) along a randomly selected line $\sL^*$ (dotted).
The boundary points (circles),
near center points (triangles), and
near boundary points (diamonds) are also shown.

\section{Computational setup and results}\label{sec:results}

The sampling method in Section~\ref{sec:sampling}
utilizes domain knowledge about the location of the boundary
to provide carefully chosen
sample points to guide the learning of the boundary
which is demonstrated in the following four examples:
two warm-up examples utilizing a quadratic
and cubic followed by two examples involving
the Kuramoto model \cite{acebron2005kuramoto,dorfler2014synchronization,strogatz2000kuramoto}.
The data sets 
for training and testing for these examples were
generated using the sampling scheme and are summarized in
Table~\ref{tab:dataSet_sizes}.  The computational time for developing the data sets was approximately 6 hours and 5 days for the 3-oscillator and 4-oscillator Kuramoto models, respectively, when computed using a single core of a 2.4 GHz AMD Opteron processor.

\begin{table}[h]
\scriptsize
  \centering
    \begin{tabular}{l|c|c|c|c}
    \toprule
% \multicolumn{5}{c}{Number of Data Points} \\
         & \multicolumn{1}{l|}{Quadratic} & \multicolumn{1}{l|}{Cubic} & \multicolumn{1}{l|}{Kuramoto $N=3$} & \multicolumn{1}{l}{Kuramoto $N=4$} \\
    \midrule
    Uniform & 10,000     & 10,000 & 972 & 8,995 \\
        \midrule
    Uniform (large) & \textbf{---} & \textbf{---} & 8000 & \textbf{---} \\
    \midrule
    
    NearBoundary & 12,934     & 12,022     & 5,192 & 54,040 \\
    \midrule
    NearBoundary+NearCenter & 25,860     & 22,036     & 8,440 & 78,823 \\
    \bottomrule
    \end{tabular}%
    \caption{Number of data points in each data set used for training/testing for each example.} %(quadratic, cubic, 3-oscillators Kuramoto, 4-oscillators Kuramoto).} 
  \label{tab:dataSet_sizes}%
\end{table}%

With these data sets, the computational
setup for using the one nearest neighbor (1-NN) and SVC-based classification was based on
\textit{KNeighborsClassifier} and \textit{svm.SVC} in {\tt SciKit-Learn} \cite{scikit-learn}, respectively, and
performed on a laptop with a 2.50 GHz Intel processor 
and 12 GB RAM.
Additionally, a feedforward network 
was utilized with computations performed
on a laptop with a six-core Intel i7 2.60 GHz processor, 
32~GB RAM, and Nvidia Quadro P2000 GPU with 4 GB of video RAM.  
The code was implemented in {\tt PyTorch} \cite{NEURIPS2019_9015}
leveraging CUDA acceleration.  Multi-layer, fully connected feedforward networks with ReLU activation functions
\cite{Hahnloser:2003:PFS:762330.762336} 
were used.  A loss function based on multi-class cross-entropy without regularization was optimized during the learning process utilizing an adaptive~learning~rate~scheme.

We note that the problems in this paper use relatively small data sets, the largest being 13.7 MB for the 4-oscillator Kuramoto model.  In all instances, the classification of the test data points using the 1-NN models, SVCs, and feedforward neural networks had a computational time of under 1 second. 
However, the time to train the SVCs and feedforward neural networks took much longer, ranging from 
seconds to hours and minutes to days, respectively.  Due to this computational expense of training, the 1-NN methods were found to be more efficient for the examples covered here.

\subsection{Quadratic}\label{sec:quadratic}

\begin{figure}[!t]
    \centering
    $\begin{array}{ccc}    
    \includegraphics[scale=0.12]{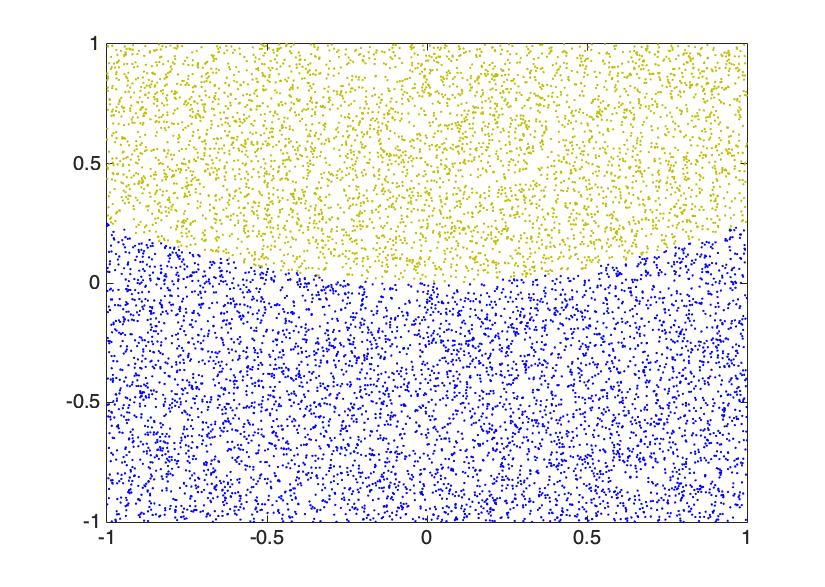} &\includegraphics[scale=0.12]{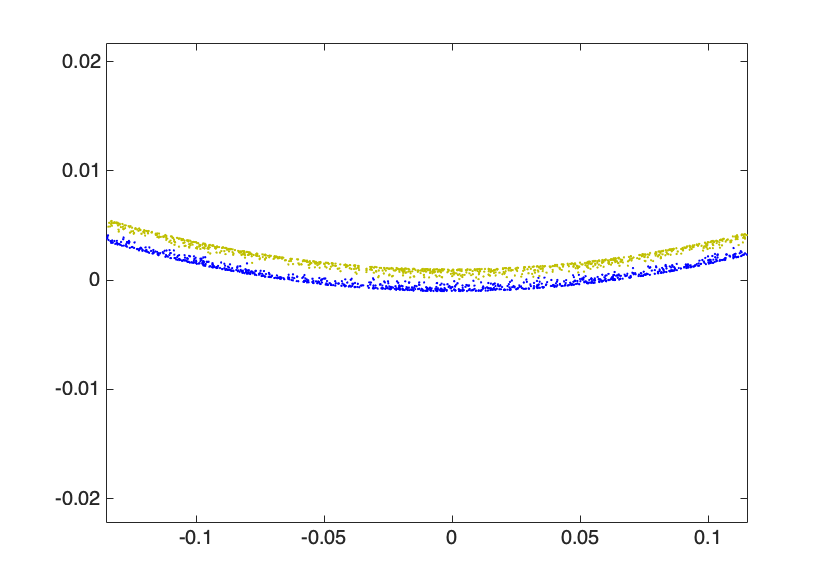}
&   \includegraphics[height = 1.02in, width = 1.4in]{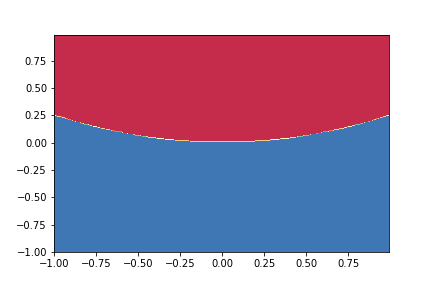} \\
    {\rm (a)} & {\rm (b)} & {\rm (c)}
    \end{array}$
    \caption{(a) Uniform random sampled data, (b) near boundary data, and (c) decision boundary from neural network trained on data from~(b) for $f(x;b,c) = x^2 + bx + c$. The blue region has 2 real solutions while the gold and red regions have 0 real solutions.}
    \label{fig:quadratic}
\end{figure}

Consider the quadratic 
$f(x;b,c) = x^2 + bx + c = 0$ 
with parameters $b$ and $c$.  This toy
system provides a demonstration of the method
restricting the parameter space to $[-1,1]^2$.
Of course, the boundary between $f$ having
$2$ real solutions and $0$ real solutions
is defined by $b^2 - 4c = 0$.
Figure~\ref{fig:quadratic}(a) plots
uniformly selected data in $[-1,1]^2$
with Figure~\ref{fig:quadratic}(b) 
showing the near boundary~data.

Table~\ref{tab:knn-quadratic} summarizes
the performance of the different classifiers on various training and testing
data sets. The results in Table~\ref{tab:knn-quadratic} were obtained with a traditional 1-NN classifier, a two-class SVC classifier with a radial basis function (RBF) kernel and inverse regularization coefficient $c = 10^6$, and a feedforward, fully connected neural network with three hidden layers each with 20 neurons. We employed $\tanh$ as the activation function for the neurons and used a 2-neuron softmax layer as the output layer. A binary cross-entropy loss without regularization was used to train the network and implemented a variable learning rate scheme. Once trained, testing data was used where each of the data points was fed to the network and the classification decision recorded. 
Figure~\ref{fig:quadratic}(c) illustrates the decision boundary learned with training data shown in Figure~\ref{fig:quadratic}(b). This plot was obtained by densely and uniformly sampling the parameter region $[-1,1]^2$, feeding the resulting samples to the trained network, and color-coding the response of the network for each of the input values in the densely sampled region. The results indicate that including sample data points near the boundary for training produces highly accurate classification results. As points near the boundary are to be classified, the performance declines when training 
with uniform data particularly with the 1-NN classifier. We attribute the relative robustness of the SVC and neural network classifiers to the sampling method via inductive bias \cite{Mitchell80} as both of those classifiers tend to learn continuous classification boundaries. In contrast, the 1-NN classifier tends to inherently overfit the training data which often results in boundaries that are not smooth.

\begin{table}[!t]
\scriptsize
  \centering
    \begin{tabular}{l|c|c|c}
    \toprule
     & \multicolumn{3}{c}{Testing Data} \\
\cmidrule{2-4} 
\multicolumn{1}{c|}{Training Data}
& \multicolumn{1}{c|}{Uniform} & \multicolumn{1}{c|}{NearBoundary} & \multicolumn{1}{l}{NearBoundary+NearCenter} \\
    \midrule
    Uniform & 1 ~/~ 1 ~/~ 1    & 0.5392 ~/~ 0.9542 ~/~ 0.9559 & 0.7678 ~/~ 0.9770 ~/~ 0.9779\\
    \midrule
    NearBoundary & 0.9999 ~/~ 1 ~/~ 1 & 1 ~/~ 1 ~/~ 1    & 1 ~/~ 1 ~/~  1 \\
    \midrule
    NearBoundary & \multirow{2}{8em}{\hspace{0.8mm}0.9999 ~/~ 1 ~/~ 1} & \multirow{2}{6em}{\hspace{1.2mm}1 ~/~ 1 ~/~ 1}  & \multirow{2}{6em}{\hspace{1.2mm}1 ~/~ 1 ~/~ 1}\\
    +NearCenter  & & & \\
    \bottomrule
    \end{tabular}%
      \caption{Accuracy rate for various testing data sets using various training data sets 
      for the 1-NN method  ~/~ SVC ~/~ feedforward neural network on $f(x;b,c)=x^2+bx+c$.}
  \label{tab:knn-quadratic}
\end{table}%

\subsection{Cubic}\label{sec:cubic}

\begin{figure}[!t]
    \centering
    $\begin{array}{ccc}   
    \includegraphics[scale=0.12]{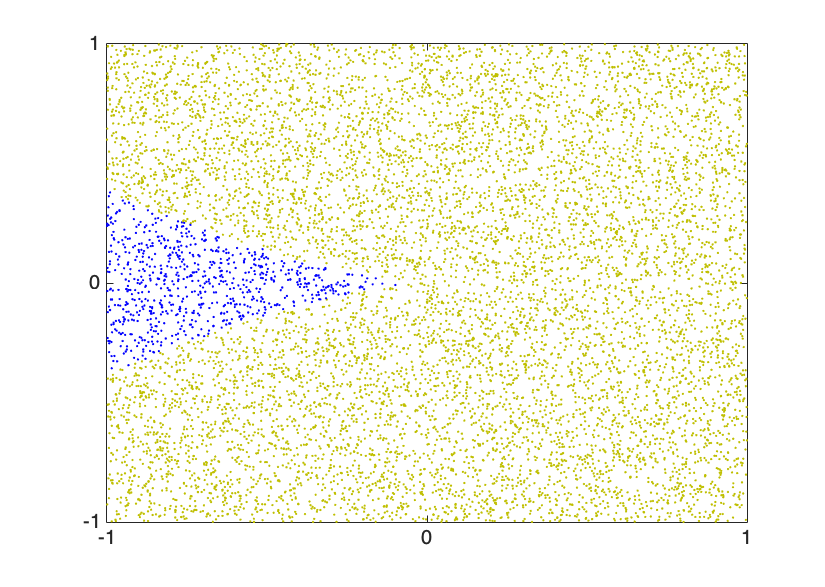} & \includegraphics[scale=0.12]{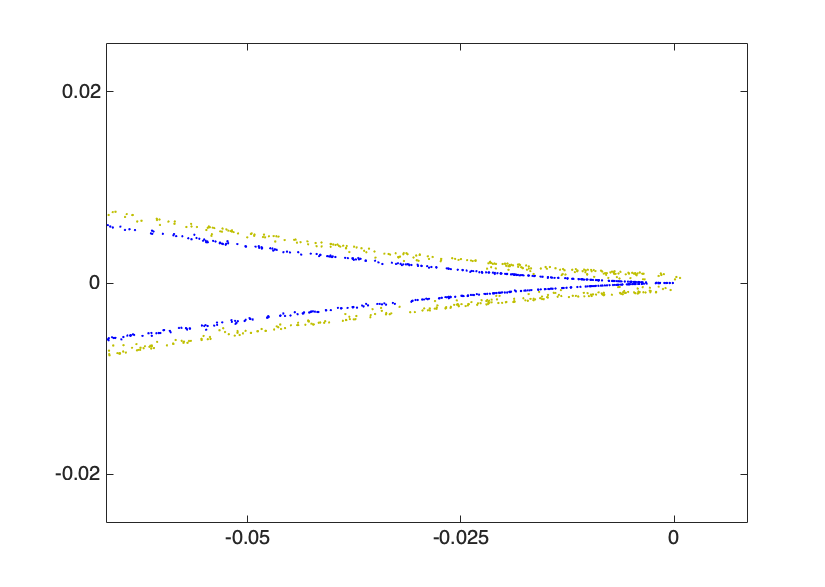}&
    \includegraphics[height = 1.02in, width = 1.4in]{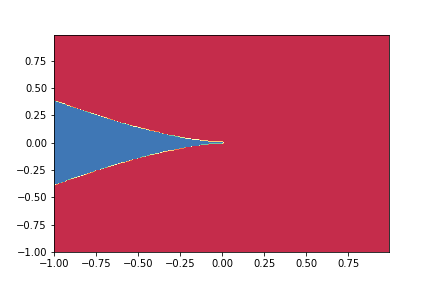}\\ %\includegraphics[scale=0.257]{CubicPaper.png}\\
    {\rm (a)} & {\rm (b)} & {\rm (c)}\end{array}$
    \caption{(a) Uniform random sampled data, (b) near boundary data near the cusp, and (c) decision boundary from neural network trained on data from~(b)
    for $f(x;b,c) = x^3 + bx + c$. The blue region has 3 real solutions while the gold and red regions have 1 real solution.}
    \label{fig:cubic}
\end{figure}

Since the real discriminant locus
for the quadratic in Section~\ref{sec:quadratic}
was smooth, we increase the degree to have
a cusp on the boundary.
In particular, we consider
the cubic $f(x;b,c) = x^3+bx+c=0$.
The boundary between $f$ having $3$ real solutions
and $1$ real solution is
defined by $4 b^3 + 27 c^2 = 0$
which has a cusp at the origin.
Figure~\ref{fig:cubic}(a) plots uniformly
selected data in $[-1,1]^2$ with
Figure~\ref{fig:quadratic}(b) showing
the near boundary data zoomed in near the cusp.

Table~\ref{tab:knn-cubic} summarizes the results obtained by the same classifiers used in Section~\ref{sec:quadratic}. As previously observed, including sample point data near the boundary for training yields higher accuracy.  When uniform data is used for training, the accuracy declines when boundary data is included in the testing data set which is particularly evident for the 1-NN classifier. Unlike competing methods, the SVC fails to fully separate the training data
 due to the somewhat limited capacity of the method which, in turn, bolsters generalization 
 capabilities in the uniformly sampled data case. Figure~\ref{fig:cubic}(c) illustrates the boundary learned by the neural network trained on the data from Figure~\ref{fig:cubic}(b).

\begin{table}[!t]
\scriptsize
  \centering
    \begin{tabular}{l|c|c|c}
    \toprule
     & \multicolumn{3}{c}{Test Data} \\
\cmidrule{2-4} \multicolumn{1}{c|}{Training Data}        & \multicolumn{1}{c|}{Uniform} & \multicolumn{1}{c|}{NearBoundary} & \multicolumn{1}{l}{NearBoundary+NearCenter} \\
    \midrule
    Uniform & 1 ~/~ 1 ~/~ 1    & 0.5259 ~/~ 0.8614 ~/~ 0.7848 & 0.7259 ~/~ 0.9116 ~/~ 0.8689 \\
    \midrule
    NearBoundary & 0.9999 ~/~ 1 ~/~ 1 & 1 ~/~ 0.9931 ~/~ 1    & 1 ~/~ 0.9951 ~/~ 1\\
    \midrule
    NearBoundary & \multirow{2}{8em}{\hspace{0.8mm}0.9999 ~/~ 1 ~/~ 1} & \multirow{2}{6em}{\hspace{1.2mm}1 ~/~ 1 ~/~ 1}    & \multirow{2}{8em}{\hspace{0.8mm}1 ~/~ 0.9912 ~/~ 1}\\
    +NearCenter & & &  \\
    \bottomrule
    \end{tabular}%
      \caption{Accuracy rate for various testing data sets using various training data sets 
      for the 1-NN method ~/~ SVC ~/~ feedforward neural network on $f(x;b,c) = x^3 + bx + c$.}
  \label{tab:knn-cubic}%
\end{table}%

\subsection{Kuramoto Model}\label{sec:kuramoto}

\begin{figure}[!b]
    \centering
    $\begin{array}{cccc}    
    \includegraphics[scale=0.11]{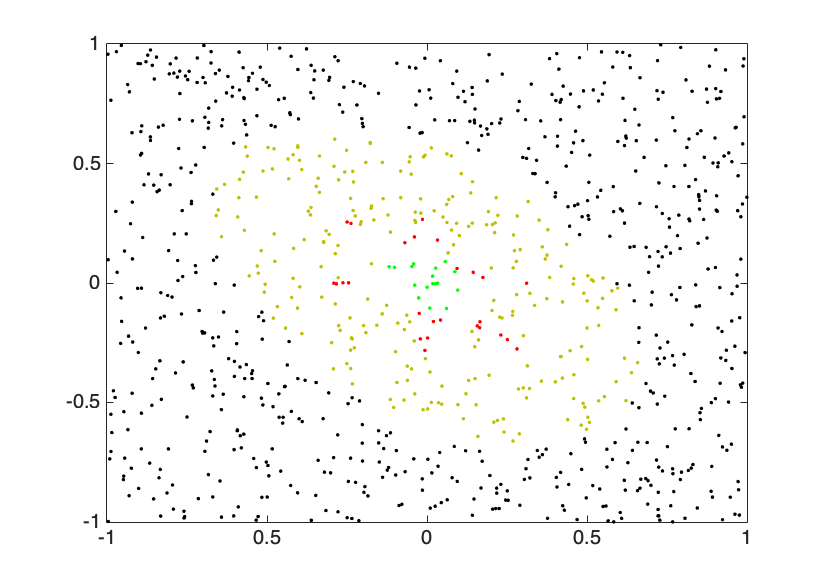}
    & \includegraphics[scale=0.11]{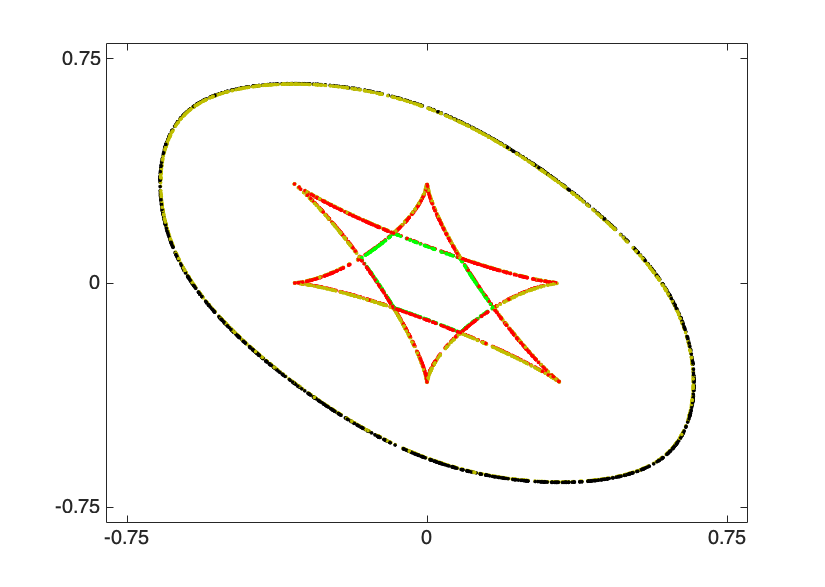}
%    \hspace{0.09in} (a) \hspace{1.85in} (b)\hspace{2.5in} \\
&    \includegraphics[scale=0.11]{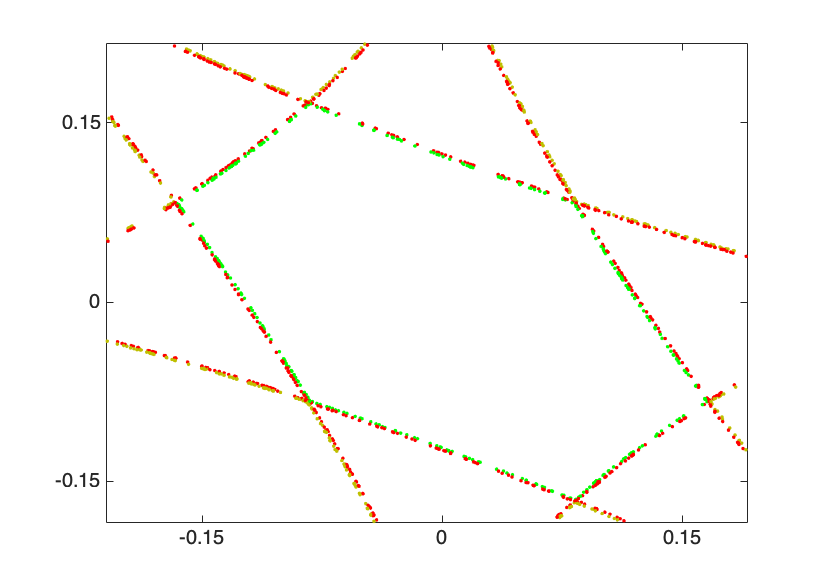} &
    \includegraphics[height = 0.96in, width = 1.25in]{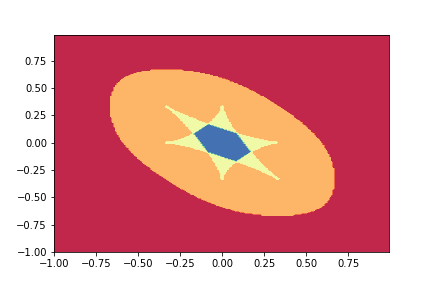} \\
{\rm (a)} & {\rm (b)} & {\rm (c)} & {\rm (d)} 
\end{array}$
    \caption{For the 3-oscillator Kuramoto model:
    (a) uniformly selected parameter values, (b) data perturbed from the boundary, (c) a zoomed view of data perturbed from the boundary, and (d) decision boundary from neural network trained on data from (b). For (d), the regions are colored based on the number of real solutions: red = 0, orange = 2, yellow = 4, and blue = 6.  These same regions are part of (a), (b), and (c) with slightly different colors.}
    \label{fig:kuramoto3}
\end{figure}

The Kuramoto model \cite{acebron2005kuramoto,dorfler2014synchronization,strogatz2000kuramoto} 
is a popular model to study synchronization phenomena observed in systems consisting of $N$ 
coupled oscillators. 
We aim to learn the number of equilibria 
as a function of the parameters $\omega\in\bR^N$
which are the natural frequencies of the $N$ oscillators.  
The system can be simplified by noting
that the sum of the natural
frequencies must be zero to have equilibria
and has rotational symmetry.  The resulting 
parameterized polynomial system 
has $2(N-1)$ polynomials and variables
with $N-1$ parameters:

\begin{equation}
\footnotesize
\begin{split}
F(c_1,s_1,\dots,c_{N-1},s_{N-1};
\omega_1,\dots,\omega_{N-1}) \\ 
& \hspace{-25mm} = \left[\begin{array}{cc}
\displaystyle \omega_i - \dfrac{1}{N}\sum_{j=1}^N(s_i c_j - s_j c_i) & \multirow{2}{*}{$i=1,\dots,N-1$} \\[0.15in]
c_i^2 + s_i^2 - 1\end{array}\right] = 0.
\end{split}
\label{eq:kuramoto_poly}
\end{equation}
Moreover, if $\omega_i\notin\left[-\frac{N-1}{N},\frac{N-1}{N}\right]$,
then Eq.~\eqref{eq:kuramoto_poly} can have no real solutions
so that the parameter space is naturally restricted
to a compact subset of $\bR^{N-1}$.
Furthermore, the number of real solutions
is invariant under permutations of
the parameters.  In particular, 
we do not label the axes in
Figures~\ref{fig:kuramoto3}
and~\ref{fig:kuramoto4} since 
equivalent pictures hold for any labelling.

For generic parameter values, 
Eq.~\eqref{eq:kuramoto_poly} has
$2^{N}-2$ solutions \cite[Thm.~4.3]{coss2018locating}.
There are a maximum
of $6$ isolated real solutions for $N =3$
and there are parameters for any possible even
number of solutions, e.g., see Figure~\ref{fig:kuramoto3}(d).
For $N = 4$, it was conjectured in
\cite{xin2016analytical} to have a maximum
of~$10$ isolated real solutions
by scanning over of a grid of the parameter space.
This conjecture was proven 
to be correct in \cite[Thm.~8.1]{Harris2020}. 
However, a complete characterization of the
parameter space based
on the number of real solutions
for the $N=4$ case has proved to be a particularly difficult problem 
for traditional methods such as comprehensive Gr\"obner basis, cylindrical algebraic decomposition,
and homotopy continuation.
We note that the complex discriminant locus
for the $N=3$ and $N=4$
cases is a curve of degree $12$
and surface of degree $48$,
respectively, and both have
singularities.
In particular, the $N=4$ Kuramoto model example highlights how the 
proposed method can be used 
to understand the parameter space
based on the number of real solutions even when
the real discriminant locus
contains a 
positive-dimensional set of 
singularities
in a reasonable time 
frame.

\subsubsection{3-oscillators}  

For $N =3$, we consider $(\omega_1,\omega_2)\in[-1,1]^2$. The outcome of a uniform sampling process is provided in  Figure~\ref{fig:kuramoto3}(a),
near boundary data points in Figure~\ref{fig:kuramoto3}(b), and a zoomed in version of near boundary points in Figure~\ref{fig:kuramoto3}(c). Table~\ref{tab:knn-k3} includes results achieved by the different classifiers on this data. For the 1-NN classifier, and due to the increasing difference in size of the data sets, tests were completed to determine whether training with a much smaller data set and testing with a data set on the order of ten times larger impacted the accuracy results.  To achieve this, the original uniform data set of approximately 1,000 data points as well as a uniform data set of 8,000 data points were used for training while data sets of approximately 1,000 (Uniform), 5,000 (NearBoundary), and 8,000 (NearBoundary+NearCenter) were used for testing.
As summarized in Table~\ref{tab:knn-k3}, the accuracy does not drastically change when the size of data sets is comparable.  Most importantly, it does not change the conclusion that including near boundary data in the training data set yields highest accuracy across all testing data sets.
A multi-class SVC with RBF kernel and inverse regularization parameter $c = 10^{10}$ was implemented. A feedforward, fully connected neural network with five hidden layers each with 20 neurons was used. We employed the ReLU activation function for the neurons in the hidden layers and used a 4-neuron softmax layer as the output layer since this is a 4-class classification task corresponding to $0$, $2$, $4$, and $6$ real solutions. In this case, the limited capacity of SVCs prevented 
fully learning the decision boundary while the competing methods perform similarly. As before, performance is lackluster on boundary data in algorithms trained only using uniform data.

\begin{table}[!t]
\scriptsize
  \centering
    \begin{tabular}{l|c|c|c}
    \toprule
    & \multicolumn{3}{c}{Test Data} \\
\cmidrule{2-4} \multicolumn{1}{c|}{Training Data}        & \multicolumn{1}{c|}{Uniform} & \multicolumn{1}{c|}{NearBoundary} & \multicolumn{1}{l}{NearBoundary+NearCenter} \\
    \midrule
    Uniform & 1 ~/~ 0.9856 ~/~ 1 & 0.4921 ~/~ 0.4962 ~/~ 0.5027 & 0.6525 ~/~ 0.6381 ~/~ 0.6554 \\
        \midrule
    Uniform (large) & \multirow{2}{8em}{\hspace{11mm}1} & \multirow{2}{8em}{\hspace{7.8mm}0.5306} & \multirow{2}{8em}{\hspace{7.8mm}0.6973}\\
    using 1-NN  & & & \\
    \midrule
    
    NearBoundary & 1 ~/~ 0.9804 ~/~ 1   & 1 ~/~ 0.7867 ~/~ 1    & 0.9985 ~/~ 0.8551 ~/~ 0.9861 \\
    \midrule
    NearBoundary & \multirow{2}{8em}{\hspace{0.8mm}1 ~/~ 0.8533 ~/~ 1}   & \multirow{2}{8em}{\hspace{0.8mm}1 ~/~ 0.7736 ~/~ 1}    & \multirow{2}{8em}{\hspace{0.8mm}1 ~/~ 0.9928 ~/~ 1}\\
    +NearCenter & & & \\
    \bottomrule
    \end{tabular}%
      \caption{Accuracy rate for various testing data sets using various training data sets 
      for the 1-NN method ~/~ SVC ~/~ feedforward neural network on the 3-oscillator Kuramoto model.} 
  \label{tab:knn-k3}%
\end{table}%

\subsubsection{4-oscillators}

\begin{figure}[!t]
    \centering
    $\begin{array}{cc}    
    \includegraphics[scale=0.155]{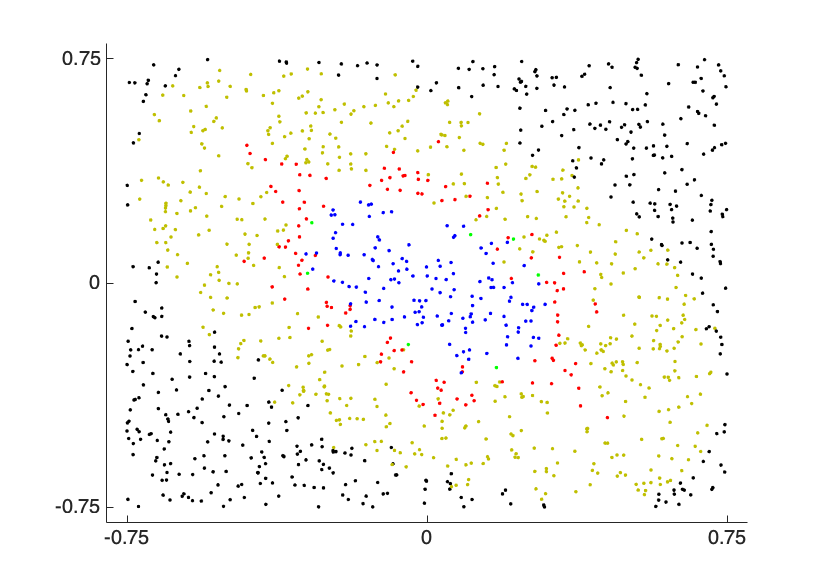} & \includegraphics[scale=0.19]{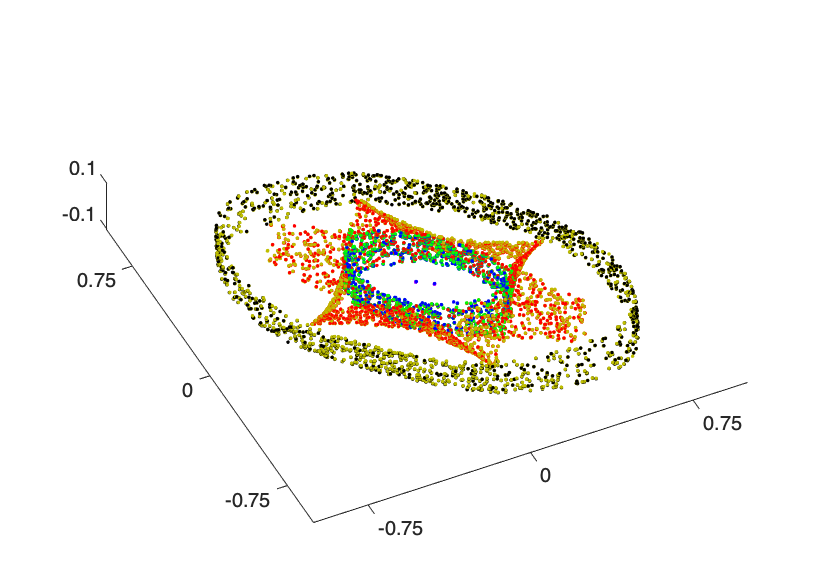}\\
{\rm (a)} & {\rm (b)} \end{array}$
    \caption{For the 4-oscillator Kuramoto model:
    (a) uniformly selected parameter values on a 2D slice,
    and (b) some of the data perturbed from the boundary colored based on the number of real solutions: black = 0, gold = 2, red = 4, green = 6, blue = 8, and magenta = 10.}
    \label{fig:kuramoto4}
\end{figure}

Similar computations were performed
on the $4$-oscillator Kuramoto model,
which has a three-dimensional parameter space. Following the theoretical bounds,
we only considered sample points in $[-3/4,3/4]^3$.
Figure~\ref{fig:kuramoto4}(a) shows
a two-dimensional slice of the parameter space
using uniformly selected points, while Figure~\ref{fig:kuramoto4}(b) 
illustrates some of the near boundary data.
Table~\ref{tab:knn-k4} summarizes the results
when using the classifiers from the previous section.

In our experiment, it became apparent that the 
neural network was unable to fully separate the data samples with 
correct labels for some of the near boundary points.
We hypothesize that, although learning converged, it likely reached a local minimum  in the optimization landscape.   As the dimensionality of the data and the number of training data points grow,  the complexity of the optimization landscape increases  which makes it less likely to reach the global minimum or at least one that is truly optimal. This scenario is worsened by the absence of a regularization term where it is empirically known~\cite{mehta2018} 
that the number of local minima in the optimization landscape of a network decreases as stronger regularization is enforced.

\begin{table}[!t]
\scriptsize
  \centering
    \begin{tabular}{l|c|c|c}
    \toprule
    & \multicolumn{3}{c}{Test Data} \\
\cmidrule{2-4} \multicolumn{1}{c|}{Training Data} 
         & \multicolumn{1}{c|}{Uniform} & \multicolumn{1}{c|}{NearBoundary} & \multicolumn{1}{l}{NearBoundary+NearCenter} \\
    \midrule
    Uniform & 1 ~/~ 0.9901     & 0.4630 ~/~ 0.5556 & 0.5911 ~/~ 0.6539 \\
    \midrule
    NearBoundary & 0.9782 ~/~ 0.9730 & 1~/~ 0.7505     & 0.9901 ~/~ 0.8059 \\
    \midrule
    NearBoundary+NearCenter & 0.9869 ~/~ 0.9901 & 1 ~/~ 0.7537    & 1 ~/~ 0.8190 \\
    \bottomrule
    \end{tabular}%
      \caption{Accuracy rate for various testing data sets using various training data sets 
      for the 1-NN method ~/~ SVC on the 4-oscillator Kuramoto model.} 
  \label{tab:knn-k4}%
\end{table}%

\section{Real parameter homotopy leveraging learned boundaries}\label{sec:realHomotopy}

The examples presented in Section~\ref{sec:results}
show that machine learning techniques
coupled with the sampling scheme
from Section~\ref{sec:sampling}
produce accurate results for classifying,
i.e., predicting the number
of real solutions, over the parameter space.
Often in science and engineering 
applications, one is not only 
interested in the number of real solutions,
but actually computing the real solutions.
Typically, for these applications, the number
of real solutions is significantly smaller
than the number of complex solutions,
so developing a parameter homotopy that 
only tracks real solution paths 
can drastically reduce the computational time.
The key to developing such a real parameter homotopy
is to track along a segment in the parameter
space which does not intersect the real discriminant locus.
Thus, after learning, one can develop
a robust and efficient real parameter homotopy setup
as follows that we demonstrate on 
the 3-oscillator and 4-oscillator Kuramoto model.

\begin{figure}[!t]
    \centering
    \includegraphics[scale=1.2]{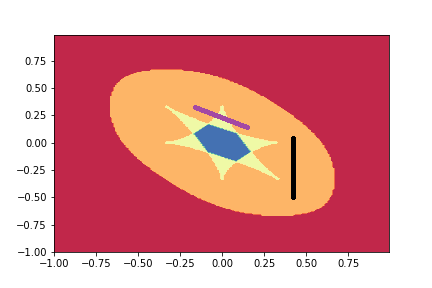}
    \caption{Illustration of
two segments added to Figure~\ref{fig:kuramoto3}(d),
one (black) which is guaranteed to succeed while the other (purple) may fail since it intersects
the real discriminant locus. See the caption of Figure~\ref{fig:kuramoto3} for a description of the color convention used.}
    \label{fig:RealHomotopy}
\end{figure}

Given a real parameter $p\in\bR^k$, the 
real parameter homotopy method uses
the nearest neighbor method to select
the closest parameter point $p^*$ to $p$ in
the sampled (training) data set.  
Since the real solutions for $f(x;p^*)=0$
have already been computed, one only 
tracks the solutions paths starting at real solutions
for the homotopy defined by
$$H(x,t) = f(x;t\cdot p^* + (1-t)\cdot p) = 0,$$
which is simply Eq.~\eqref{eq:ParameterHomotopy} with $\gamma=1$.
Therefore, if the line segment $[p^*,p]$
does not intersect the real discriminant locus,
then there is a bijection between the real solutions
of $f(x;p) = 0$ and $f(x;p^*)=0$,
and every real solution path of $H=0$ is smooth
for $t\in[0,1]$.
Using sample points via the sampling scheme
in Section~\ref{sec:sampling} on either side of
the boundary aims to increase the chance
the segment between~$p$ and 
the nearest sample point $p^*$ is
contained in the same region and thus this real parameter
homotopy method succeeds.  Figure~\ref{fig:RealHomotopy}
is Figure~\ref{fig:kuramoto3}(d) 
with two added segments.  
One segment (black) is within the same region 
so that the real parameter homotopy method would succeed.
Although the other segment (purple) has endpoints with the 
same region, there is no guarantee of success
since it intersects the real discriminant locus.

\subsection{3-oscillators}

As an illustration, consider the 
3-oscillator Kuramoto model.  
Since, from Section~\ref{sec:kuramoto},
the generic number of complex solutions is $6$,
one of course can easily track all $6$ complex solution
paths using a classical parameter homotopy in Eq.~\eqref{eq:ParameterHomotopy}.  
In our experiment, 
using a single core of a
2.4 GHz AMD Opteron Processor,
this took on average $1.33$ seconds.
Nonetheless, we utilize this as a test
case to show some improvement as well as
analyzing the success rate
which was determined by comparing using a classical
parameter homotopy with this
machine learning assisted real parameter homotopy.
Table~\ref{tab:K3_real_parameter_homotopy} 
shows that, on average, the real parameter
homotopy took less than $0.1$ seconds
and was successful on every randomly selected
parameter value tested.
One reason for the order of magnitude 
reduction in computational time is that,
by selecting the closest parameter value,
the homotopy solution paths are much shorter
and thus faster to track.

\begin{table}[!t]
\scriptsize
    \centering
    \begin{tabular}{c|c|c|c}
    \toprule
    Number of data points     & Number of paths & Average time (in seconds) & Success rate \\
    \midrule
        249 & 2 & 0.077 & $100\%$\\
         \midrule
         26& 4 & 0.081 & $100\%$\\
         \midrule
         17 & 6 & 0.086 &$100\%$ \\
         \bottomrule
    \end{tabular}
    \caption{Average computation time for finding all real roots for the 3-oscillator Kuramoto model using a machine-learning-assisted real parameter homotopy method.}
    \label{tab:K3_real_parameter_homotopy}
\end{table}

\subsection{4-oscillators}

Following a similar setup, we also applied
the method to the 4-oscillator Kuramoto model. 
In this case, the generic number of complex
solutions is $14$, but the maximum number of real
solutions is $10$ showing that there will always
be wasted computational effort when computing
the real solutions using a classical parameter homotopy.
In our experiment, the average time
for tracking the 14 complex solution paths 
using a classical parameter homotopy was $3.40$ seconds.
Table~\ref{tab:K4_2_real_parameter_homotopy}
summarizes the results that again show
over an order of magnitude reduction
in computational time with a success rate
in accordance with the classification accuracy 
in Table~\ref{tab:knn-k4}.

\begin{table}[!t]
\scriptsize
    \centering
    \begin{tabular}{c|c|c|c}
    \toprule
    Number of data points     & Number of paths & Average time (in seconds) & Success rate \\
    \midrule
        30504  & 2 & 0.114 & $98.2\%$\\
         \midrule
         17088& 4 & 0.121 & $98.8\%$\\
         \midrule
         9041 & 6 & 0.126 &$99.2\%$ \\
           \midrule
         4383 & 8 & 0.128 &$98.0\%$ \\
           \midrule
         345 & 10 & 0.132 &$100\%$ \\
         \bottomrule
    \end{tabular}
    \caption{The average computation time for finding all real roots for the 4-oscillator Kuramoto model using a machine learning assisted real parameter homotopy method.}
    \label{tab:K4_2_real_parameter_homotopy}
\end{table}

\section{Outlook and conclusions}\label{sec:conclusion}

This paper provides a novel viewpoint
on the mathematical problem of identifying
the boundaries, called the real discriminant
locus, of the parameter space
that separate the regions corresponding
to different number of real solutions
to a parameterized polynomial system.
Although there is a discriminant polynomial
which vanishes on the real discriminant locus,
it can be difficult to compute, facilitating
the need to numerically approximate it.
Our approach is based on the correspondence
between the real discriminant locus
and decision boundaries of
a supervised classification problem in machine learning.
By utilizing domain knowledge from numerical algebraic
geometry, we developed a sampling strategy for selecting
points near the boundary to assist the machine learning techniques in providing an accurate approximation
of the boundary.  
With a parameter homotopy, one is able to accurately
label the data so that there is no noise
in the data.  Hence, no regularization techniques
need to be utilized, which would have
forced the algorithm to strictly learn only 
smooth boundaries, which is important
since singularities often arise as illustrated 
in Section~\ref{sec:results}.

One challenge with using deep networks
to learn a real discriminant locus
is how to properly select the number of
layers and neurons within each layer
needed to develop an accurate approximation.
We utilized hyperparameter optimization
to search for reasonable choices
along with stochastic gradient descent methods
to determine weights to fit the data.
Another challenge is the presence
of singularities which
seem to make training more difficult for deep networks.
Therefore, these type of problems
provide a unique benchmarking opportunity for multi-class machine learning algorithms as the ground truth 
regarding both labels and classification 
boundaries can be explicitly computed 
for some examples, such as univariate polynomials
as in Sections~\ref{sec:quadratic} and~\ref{sec:cubic}.
We overcome some of these difficulties 
by developing a sampling scheme
that produces significantly more points near the boundaries than in other areas of the parameter space
so that one is able to quickly obtain an accurate
approximation of the real discriminant locus.

When deep networks can take an inordinate
amount of time to train, one can utilize
local approximation methods such as
$k$-nearest neighbor classification algorithm.
In fact, as shown in Theorem~\ref{thm:1NN},
no classifier can outperform the $1$-nearest neighbor
classification algorithm provided that the 
parameter space is sampled densely enough.
The examples in Section~\ref{sec:results}
show that deep networks can be useful and comparable to the 1-NN methods.  However, the data confirms the effectiveness of the 1-NN methods, especially in the case of the $N=4$ Kuramoto model when the deep network did not converge to the global minimum.

Although our proposed sampling method 
can be viewed as active learning, 
one can also employ a more explicit active learning 
approach where an algorithm interactively queries 
the parameter space and samples more densely near 
singularities such as cusps and other difficult regions. 
One could also attempt to first construct an algorithm to remove $\epsilon$ neighborhoods surrounding all singularities,
learn the remaining parameter space
and real discriminant locus,
and then take $\epsilon \rightarrow 0$. These approaches will be explored in the future. 

The curse of dimensionality hampers most computational methods, including machine learning. 
In computational algebraic geometry, the actual dimension where problems become intractable
is strictly problem specific.  When identifying the real discriminant locus for
parameterized polynomial systems, this brings in its own difficulties 
that are different from other applications such as those in computer vision and natural language processing where the curse of dimensionality may kick in at much larger dimensions. 
The previous best works \cite{chandra2017locating, harrington2016decomposing} proposed
an approach based on homotopy continuation which could analyze the $N=3$ Kuramoto model, while the $N=4$ case was still out of reach. 
In the present work, we have now broken this ceiling with the combination of machine learning and homotopy continuation.

A real parameter homotopy method
that tracks only real solution paths was developed in Section~\ref{sec:realHomotopy}
as an application of learning the real discriminant
locus.  Even for relatively small problems, 
this method reduced the computational
time by over an order of magnitude.
After generating sample data ``offline,''
this method is easy to implement
in an ``online'' solver which could drastically
improve the computation of real solutions.
With proper adjustments,
this method is extensible to other situations
involving rational, exponential,
logarithmic, trigonometric, and piecewise~functions.

\section*{Acknowledgments}
The paper is a result of exploratory and fundamental research, and statements made in it are Dhagash Mehta's and his co-authors' personal views which do not represent The Vanguard Group's views. The authors thank Martin Pol\'{a}\v{c}ek for insightful comments.  JDH was supported in part by NSF grant CCF-1812746 and ONR N00014-16-1-2722. MHR was supported in part by Schmitt Leadership Fellowship in Science and Engineering and NSF grant CCF-1812746.

\bibliographystyle{abbrv}
\bibliography{bibliography_NPHC_NAG}

\end{document}